\documentclass[runningheads]{llncs}
\usepackage[T1]{fontenc}
\usepackage{layouts}

\usepackage{bm}
\usepackage{graphicx}
\usepackage{amsmath, amssymb, hyperref, amsfonts}
\usepackage{tikz}
\usetikzlibrary{calc}
\usepackage{cite}
\usepackage[section]{placeins}
\usepackage{caption}
\usepackage{subcaption}
\tikzset{neuron/.style={shape=circle, minimum size=1.6cm, inner sep=0.2, draw, font=\large}, 
  input_io/.style={shape=circle, minimum size=1.5cm, inner sep=0.2, draw, font=\large, fill=cyan!10}, 
  output_io/.style={shape=circle, minimum size=2.0cm, inner sep=0.2, draw, font=\large, fill=orange!15}, 
  dense/.style={shape=ellipse, minimum width=1.5cm, minimum height=0.8cm, inner sep=0.2, draw, font=\large}}
  \tikzset{font=\normalsize}
 \usetikzlibrary{positioning, fit, arrows.meta, shapes}
\usepackage{color}

\begin{document}
\title{Physics-Informed Long Short-Term Memory for Forecasting and Reconstruction of Chaos\thanks{This research is supported by the ERC Starting Grant No. PhyCo 949388.}}
\titlerunning{PI-LSTM for Forecasting and Reconstruction of Chaos}
%
\author{Elise Özalp\inst{1} \and Georgios Margazoglou \inst{1} \and
Luca Magri\inst{1, 2}}
\authorrunning{E. Özalp et al.}
%
\institute{Department of Aeronautics, Imperial College London, London, SW7 2AZ, UK\and
The Alan Turing Institute, London, NW1 2DB, UK \\
\email{l.magri@imperial.ac.uk}
}

%
\maketitle              
\begin{abstract}



We present the Physics-Informed Long Short-Term Memory (PI-LSTM) network to reconstruct and predict the evolution of unmeasured variables in a chaotic system. The training is constrained by a regularization term, which penalizes solutions that violate the system's governing equations. 
The network is showcased on the Lorenz-96 model, a prototypical chaotic dynamical system, for a varying number of variables to reconstruct.
First, we show the PI-LSTM architecture and explain how to constrain the differential equations, which is a non-trivial task in LSTMs. Second, the PI-LSTM is numerically evaluated in the long-term autonomous evolution to study its ergodic properties. We show that it correctly predicts the statistics of the unmeasured variables, which cannot be achieved without the physical constraint. 
Third, we compute the Lyapunov exponents of the network to infer the key stability properties of the chaotic system. For reconstruction purposes, adding the physics-informed loss qualitatively  enhances the dynamical behaviour of the network, compared to a data-driven only training. This is quantified by the agreement of the Lyapunov exponents. This work opens up new opportunities for state reconstruction and learning of the dynamics of nonlinear systems.

\end{abstract}
\section{Introduction}
Chaotic dynamics arise in a variety of disciplines such as meteorology, chemistry, economics, and engineering. Their manifestation emerges because of the exponential sensitivity to initial conditions, which makes long-term time-accurate prediction difficult. In many engineering cases, only partial information about the system's state is available, e.g., because of the computational cost or a limited number of sensors in a laboratory experiment and hence, no data is available on the unmeasured variables. Making predictions of the available observations and reconstructing the unmeasured variables is key to understanding and predicting the behaviour of dynamical systems.

Neural networks are powerful expressive tools to extract patterns from data and, once trained, they are fast and efficient at making predictions. Suitable for time series and dynamical evolutions are recurrent neural networks (RNNs) and long short-term memory networks (LSTMs), which have shown promising performance in the inference of dynamical systems with multi-scale, chaotic, or turbulent behaviour \cite{Vlachas_backprop}. The integration of neural networks with knowledge of governing physical equations has given rise to the field of physics-informed neural networks \cite{lagaris, raissi2019physics}. More specifically, physics-informed RNNs with reservoir computers have been applied successfully in the short-term reconstruction of unmeasured variables~\cite{racca2021automatic, doan_piesn}. 



In this paper, we propose a physics-informed LSTM (PI-LSTM) to make predictions of observed variables and, simultaneously, infer the unmeasured variables of a chaotic system. The quality of the prediction is evaluated by analysing the autonomous long-term evolution of the LSTM and collecting the statistics of the reconstructed variables. Crucial quantities for characterizing chaos are the Lyapunov exponents (LEs), which provide insight into an attractor's dimension and tangent space. In this paper, we extract the LEs from an LSTM, trained on a prototypical chaotic dynamical system. For state reconstruction, we show that it is necessary to embed prior knowledge in the LSTM, which is in the form of differential equations in this study. 




The paper is structured as follows. Section \ref{sec:chaos_dyn_sys} provides a brief introduction to the LEs of chaotic systems and the problem setup of unmeasured variables. The PI-LSTM is proposed in Section \ref{sec:pi_lstm}. In Section \ref{sec:L96}, we discuss the results for the Lorenz-96 system. Finally, we summarize the work and propose future direction in Section \ref{sec:conclusion}. 
\section{Chaotic dynamical systems}\label{sec:chaos_dyn_sys}

We consider a nonlinear autonomous dynamical system
\begin{equation}\label{autonom_dyn_system}
 \frac{d}{dt} \bm{y}(t) = f(\bm{y}(t)),
\end{equation}
where $\bm{y}(t) \in \mathbb{R}^N$ is the state vector of the physical system and $f:\mathbb{R}^N \to \mathbb{R}^N$ is a smooth nonlinear function. The dynamical system \eqref{autonom_dyn_system} is chaotic if infinitesimally nearby trajectories diverge at an exponential rate. This behaviour is quantified by the LEs, which measure the average rate of stretching of the trajectories in the phase space. The LEs, $\lambda_1 \geq \dots \geq \lambda_N$, provide fundamental insight into the chaoticity and geometry of an attractor. Chaotic dynamical systems have at least one positive Lyapunov exponent. In chaotic systems, the Lyapunov time $\tau_{\lambda} = \frac{1}{\lambda_1}$ defines a characteristic timescale for two nearby trajectories to separate, which gives an estimate of the system’s predictability horizon. LEs can be computed numerically by linearizing the equations of the dynamical system around a reference point, defining the tangent space, and by extracting the LEs based on the Gram-Schmidt orthogonalization procedure from the corresponding Jacobian \cite{Eckmann_Ruelle1985}.

\subsection{State reconstruction}
Let $\bm{y}(t) = [\bm{x}(t); \bm{\xi}(t)]$ be the state of a chaotic dynamical system, where $\bm{x}(t) \in \mathbb{R}^{N_x}$ are the observed variables and $\bm{\xi}(t) \in \mathbb{R}^{N_{\xi}}$ are the unmeasured variables with $N = N_x + N_{\xi}$. Specifically, let us assume that $\bm{x}(t_i)$ is measured at times $t_i = i \Delta t $ with $i=0, \dots N_{t}$ and constant time step $\Delta t$. Based on these observations, we wish to predict the full state $\bm{y}(t_i) =[\bm{x}(t_i), \bm{\xi}(t_i)]$, whilst respecting the governing equation~\eqref{autonom_dyn_system}, using the PI-LSTM.

\section{Physics-Informed Long-Short Term Memory}\label{sec:pi_lstm}

LSTMs have been successfully applied to time forecasting of dynamical systems when full observations are available \cite{Sangiorgio_2021, Vlachas_backprop}. They are characterized by a cell state $\bm{c}_{i+1} \in \mathbb{R}^{N_{h}}$ and a hidden state $\bm{h}_{i+1} \in \mathbb{R}^{N_{h}}$ that are updated at each step. 
In the case of partial observations, the states are updated using the observed variables $\bm{x}(t_i)$ as follows\\
\begin{minipage}{.5\linewidth}
  \begin{align*}
    \bm{i}_{i+1} &= \sigma \left(\bm{W}^i [\bm{x}(t_i); \bm{h}_{i}] + \bm{b}^i \right), \\
    \bm{f}_{i+1} &= \sigma \left(\bm{W}^f [\bm{x}(t_i); \bm{h}_{i}] + \bm{b}^f \right), \\
    \bm{o}_{i+1} &= \sigma \left(\bm{W}^o [\bm{x}(t_i); \bm{h}_{i}] + \bm{b}^o \right),
  \end{align*}
\end{minipage}%
\begin{minipage}{.5\linewidth}
  \begin{align*}\label{LSTM_equation*s}
    \bm{\Tilde{c}}_{i+1} &= \tanh{\left(\bm{W}^g [\bm{x}(t_i); \bm{h}_{i}] + \bm{b}^g \right) }, \nonumber\\ 
    \bm{c}_{i+1} &= \sigma \left( \bm{f}_{i+1}*\bm{c}_{i} + \bm{i}_{i+1}*\bm{\Tilde{c}}_{i+1} \right), \\
    \bm{h}_{i+1} &= \tanh{\left(\bm{c}_{i+1} \right)} * \bm{o}_{i+1}, \nonumber
  \end{align*}
\end{minipage}
where $\bm{i}_{i+1}, \bm{f}_{i+1}, \bm{o}_{i+1} \in \mathbb{R}^{N_{h}}$ are the input, forget and output gates. The matrices $\bm{W}^i, \bm{W}^f, \bm{W}^o, \bm{W}^g \in \mathbb{R}^{N_{h} \times (N_x+N_h)}$ are the corresponding weight matrices, and $\bm{b}^i, \bm{b}^f, \bm{b}^o, \bm{b}^g \in \mathbb{R}^{N_{h}}$ are the biases. 
The full prediction $\bm{\Tilde{y}}(t_{i+1})= [\bm{\Tilde{x}}(t_{i+1}), \bm{\Tilde{\xi}}(t_{i+1}) ]$ is obtained by concatenating the hidden state $\bm{h}_{i+1}$ with a dense layer
\begin{align*}
    \begin{bmatrix} \bm{\Tilde{x}}(t_{i+1}) \\ \bm{\Tilde{\xi}}(t_{i+1})  \end{bmatrix}= \bm{W}^{dense} \bm{h}_{i+1} + \bm{b}^{dense},
\end{align*}
where $\bm{W}^{dense} \in \mathbb{R}^{(N_x+N_{\xi})\times N_h}$ and $\bm{b}^{dense} \in \mathbb{R}^{N_x+N_{\xi}} $.

\begin{figure}[t]
\centering
\begin{minipage}{.5\textwidth}
  \centering
  \scalebox{0.5}{
  \begin{tikzpicture}[x=2.75cm, y=1.25cm, >=Stealth]
    \foreach \jlabel [count=\j, evaluate={\k=int(mod(\j-1,1)); \jj=int(\j-1);}]
      in {0, 1}{
        \foreach \ilabel [count=\i] in {1}
            \node [neuron, fill=gray!10, align=left] at (\j, 1-\i) (h-\i-\j){ $\mathbf{LSTM}$};   
          \node [fit=(h-1-\j) (h-1-\j), inner sep=0, draw] (b-\j){ } ;
          \node [input_io, below=of b-\j] (v-\j) {$ \bm{x}(t_{ \jlabel})$};
        \node [dense, above=of h-1-\j, align=left] (d-\j) {{{Dense}}};
          \draw [->] (v-\j) -- (b-\j);
          \draw [->] (b-\j.north) -- (d-\j) node [midway, right] {$\bm{h}_{\j}$};
          }
        \node [output_io, above=of d-1] (output1){$   \begin{bmatrix}
         \bm{\Tilde{x}}(t_{1})\\ \bm{\Tilde{\xi}}(t_{1})\end{bmatrix}
           $ };
         \node [output_io, above=of d-2] (output2){$   \begin{bmatrix}
         \bm{\Tilde{x}}(t_{2})\\ \bm{\Tilde{\xi}}(t_{2})\end{bmatrix}
           $ };
        \node [right=1.5cm of h-1-2](dots1) {\ldots};
        \node [neuron, fill=gray!10,  align=left] at (4, 0) (h-1-4){$\mathbf{LSTM}$ 
            };
        \node [fit=(h-1-4) (h-1-4), inner sep=0, draw] (b-4){} ;
        \node [input_io, below=of b-4] (v-4) {$  \bm{x}(t_{ n-1})  $};
         \node [right=1.5cm of v-2](dots2) {\ldots};
        \draw [->] (v-4) -- (b-4);
        \draw [->] (h-1-1.east) -- (h-1-2.west)node [midway, above] {$\bm{c}_1, \bm{h}_{1}$} ;
        \draw [->] (h-1-2.east) -- (dots1.west)node [midway, above] {$\bm{c}_2, \bm{h}_{2}$};
        \draw [->] (dots1.east) -- (h-1-4.west)node [midway, above] {$\bm{c}_{n-1}, \bm{h}_{n-1}$};
        \node [dense, above=of b-4, align=left] (d-4) {{{Dense}} };
        \draw [->] (b-4.north) -- (d-4.south)node [midway, right] {$\bm{h}_{n}$} ;
        \node [output_io,  above=of d-4] (output){$   \begin{bmatrix}
         \bm{\Tilde{x}}(t_{ n})\\ \bm{\Tilde{\xi}}(t_{ n})\end{bmatrix}
       $ };
        \draw [->] (d-4.north) -- (output.south);
        \draw [->] (d-1.north) -- (output1.south);
        \draw [->] (d-2.north) -- (output2.south);
     \draw[|->] (0.2,0) -- (b-1.west)node [midway, above] {$\bm{c}_0, \bm{h}_{0}$};
    \end{tikzpicture}}
    \caption{Open-loop configuration.}
    \label{fig:open_loop}
\end{minipage}%
\begin{minipage}{.5\textwidth}
\centering
    \scalebox{0.5}{
   \begin{tikzpicture}[x=2.75cm, y=1.25cm, >=Stealth]

        \node [neuron, fill=gray!10, align=left] at (4.25, 0) (h-1-4){$\mathbf{LSTM}$ 
            };
        \node [fit=(h-1-4) (h-1-4), inner sep=0, draw] (b-4){} ;
        \node [input_io, below=of h-1-4] (v-4) {$  \bm{x}(t_{ n})  $};
        \draw [->] (v-4) -- (h-1-4);
        \draw [->] (3.25, 0) -- (h-1-4.west)node [midway, above] {$\bm{c}_{n}, \bm{h}_{n}$};
        \node [dense, above=of h-1-4, align=left] (d-4) {{{Dense}} };
        \draw [->] (h-1-4.north) -- (d-4.south)node [midway, right] {$\bm{h}_{n+1}$} ;
        \node [output_io, above=of d-4] (output){$  \begin{bmatrix}
         \bm{\Tilde{x}}(t_{ n+1})\\ \bm{\Tilde{\xi}}(t_{ n+1})\end{bmatrix}
       $ };
        \draw [->] (d-4.north) -- (output.south);
        \node [neuron, fill=gray!10, align=left] at (5.5, 0) (h-1-6){$\mathbf{LSTM}$ 
            };
        \node [fit=(h-1-6) (h-1-6), inner sep=0, draw] (b-6){} ;
        
         \draw[->, rounded corners=10pt, gray, dashed] (output.east) -- ++(0.25, 0) -- ++(0, -6) -- ++(0.6, 0) -- (h-1-6.south);
        \node [input_io, fill=orange!15, below=of h-1-6] (v-6) {$  \Tilde{\bm{x}}(t_{ n+1})  $};
    \draw [->] (h-1-4.east) -- (h-1-6.west)node [midway, above] {$\bm{c}_{n+1}, \bm{h}_{n+1}$};
        \node [dense, above=of b-6, align=left] (d-6) {{{Dense}} };
        \draw [->] (b-6.north) -- (d-6.south)node [midway, right] {$\bm{h}_{n+2}$} ;
        \node [output_io, above=of d-6] (output_n){$ \begin{bmatrix}
         \bm{\Tilde{x}}(t_{ n+2})\\ \bm{\Tilde{\xi}}(t_{ n+2})\end{bmatrix}$ };
       \draw [->] (d-6.north) -- (output_n.south);
        \node [neuron, fill=gray!10, align=left] at (6.75, 0) (h-1-7){$\mathbf{LSTM}$ 
            };
        \node [fit=(h-1-7) (h-1-7), inner sep=0, draw] (b-7){} ;        
         \draw[->, rounded corners=10pt, gray, dashed] (output_n.east) -- ++(0.25, 0) -- ++(0, -6) -- ++(0.6, 0) -- (h-1-7.south);
        \node [input_io, fill=orange!15, below=of h-1-7] (v-7) {$  \Tilde{\bm{x}}(t_{ n+2})  $};
    \draw [->] (h-1-6.east) -- (h-1-7.west)node [midway, above] {$\bm{c}_{n+2}, \bm{h}_{n+2}$};
        \node [dense, above=of b-7, align=left] (d-7) {{{Dense}} };
        \draw [->] (b-7.north) -- (d-7.south)node [midway, right] {$\bm{h}_{n+3}$} ;
        \node [output_io, above=of d-7] (output_n1){$  \begin{bmatrix}
         \bm{\Tilde{x}}(t_{n+3})\\ \bm{\Tilde{\xi}}(t_{n+3})\end{bmatrix}$ };
       \draw [->] (d-7.north) -- (output_n1.south);
       
    \draw [->] (h-1-7.east) -- (7.75, 0)node [midway, above] {$\bm{c}_{n+3}, \bm{h}_{n+3}$};
            
    \end{tikzpicture}
     }
    \caption{Closed-loop configuration.}
    \label{fig:closed_loop}
\end{minipage}
\end{figure}
 LSTMs are universal approximators for an arbitrary continuous target function\cite{FUNAHASHI_nn_approx, hornik1989multilayer}; however, practically, the network's performance is dependent on the parameters, such as weights and biases, which are computed during the training phase. To train the weights and biases, a data-driven loss is defined on the observed data via the mean-squared error $
    \mathcal{L}_{dd} = \frac{1}{N_{t}} \sum_{i=1}^{N_{t}} \left(\bm{x}(t_i) - \bm{\Tilde{x}}(t_i)\right)^2$.
To constrain the network for the unmeasured dynamics, we add a penalization term 
\begin{equation*}\label{physics_loss}
        \mathcal{L}_{pi} = \frac{1}{N_{t}} \sum_{i=1}^{N_{t}} \biggl(\frac{d}{dt} \bm{\Tilde{y}}(t_{i}) - f(\bm{\Tilde{y}}(t_{i})) \biggr)^2.
\end{equation*}
This loss regularizes the network's training to provide predictions that fulfil the governing equation from \eqref{autonom_dyn_system} (up to a numerical tolerance).
For simplicity, the time derivative $\frac{d}{dt}\bm{\Tilde{y}}$ is computed using a forward difference scheme
\begin{equation*}\label{fd_for_phyiscs_loss}
    \frac{d}{dt}\bm{\Tilde{y}}(t_i) \approx \frac{ \bm{\Tilde{y}}(t_{i+1}) - \bm{\Tilde{y}}(t_{i})}{\Delta t}.
\end{equation*}
(Alternatively, the time derivative can be computed with a higher-order scheme.) Combining the data-driven loss and weighing the physics-informed loss leads to the total loss
\begin{equation}
    \mathcal{L} = \mathcal{L}_{dd}  + \alpha_{pi}\mathcal{L}_{pi}, \quad \alpha_{pi} \in \mathbb{R}^{+}, 
\end{equation}
where $\alpha_{pi}$ is a penalty hyperparameter. 
If $\alpha_{pi} = 0$, the network is not constrained by the governing equations, which is referred to as the `data-driven LSTM'.

The weights and biases are optimized by minimizing the loss $\mathcal{L}$ with the Adam optimizer~\cite{adam}. Early stopping is employed to avoid overfitting. During the training and validation, the network is in an open-loop configuration, as shown in Fig.~\ref{fig:open_loop}. After training, the network is evaluated on test data with fixed weights and biases, operating in a closed-loop configuration, as shown in Fig.~\ref{fig:closed_loop}. 
In this mode, the network predicts the observed variables and the unmeasured variables, while the observed variables are used as input for the next time step, allowing for an autonomous evolution of the LSTM. This effectively defines a dynamical system and allows for stability analysis to be performed on the LSTM.

Previous work showed that when Echo State Networks and Gated Recurrent Units are employed to learn and predict full states from chaotic systems, the LEs of the network align with those of the dynamical system \cite{georgios_stability_analysis, Vlachas_backprop, Pathak_ml_le} (to a numerical tolerance), allowing to gain valuable insight into the behaviour of the network. Thus, we analyse the LEs of the proposed PI-LSTM. 

\section{State reconstruction and Lyapunov exponents of the Lorenz-96 model}\label{sec:L96}
The Lorenz-96 model is a system of coupled ordinary differential equations that describe the large-scale behaviour of the mid-latitude atmosphere, and the transfer of a scalar atmospheric quantity \cite{lorenz96}. The explicit Lorenz-96 formulation of dynamical system~\eqref{autonom_dyn_system} is

\begin{equation}
\frac{d}{dt} y_i(t)= \left( y_{i+1}(t) - y_{i-2}(t)\right) y_{i-1}(t) - y_i(t) + F, \;\;\;\; i=1,\ldots,N,
\end{equation}
where $F$ is an external forcing. It is assumed that $y_{-1}(t) = y_{N-1}(t)$, $y_0(t) = y_N(t)$ and $y_{N+1}(t) = y_1(t)$. The state $\bm{y}(t)\in\mathbb{R}^N$ describes an atmospheric quantity in $N$ sectors of one latitude circle. 
In this study, we set $F=8$ and $N=10$, for which the system exhibits chaos with three positive LEs. Both the numerical solution and the reference LEs are computed using a first-order Euler scheme with a time step of $\Delta t= 0.01$. The largest Lyapunov exponent is $\lambda_1\approx 1.59$.
The training set consists of $N_t=20000$ points, which is equivalent to $125 \tau_{\lambda}$. Hyperparameter tuning was utilized to optimise key network parameters. In particular, the dimension of the hidden and cell state $N_h$ was selected from $\{20, 50, 100\}$ and the weighing of the physics-informed loss $\alpha_{pi}$ varied from $ 10^{-9}$ to $1$ in powers of $10$.

We deploy the PI-LSTM to reconstruct the unmeasured variables in three test cases: reconstruction of (i) $N_{\xi}=1$, (ii) $N_{\xi}=3$, (iii) $N_{\xi}=5$ unmeasured variables. We display networks with parameters (i) $N_h=100, \alpha_{pi}=0.01$, (ii) $N_h=100, \alpha_{pi}=0.01$, and (iii) $N_h=50, \alpha_{pi}=0.001$. We choose test case (i) and (ii) to highlight the capabilities of the PI-LSTM and select (iii) to demonstrate how the network behaves with further limited information. (We remark that, when the full state is available, both data-driven LSTM and PI-LSTM perform equally well in learning the long-term statistics and the LEs.) The reconstruction is based on an autonomous $1000 \tau_{\lambda}$ long trajectory in closed-loop mode. 
\begin{figure}[h]
    \centering
    \includegraphics{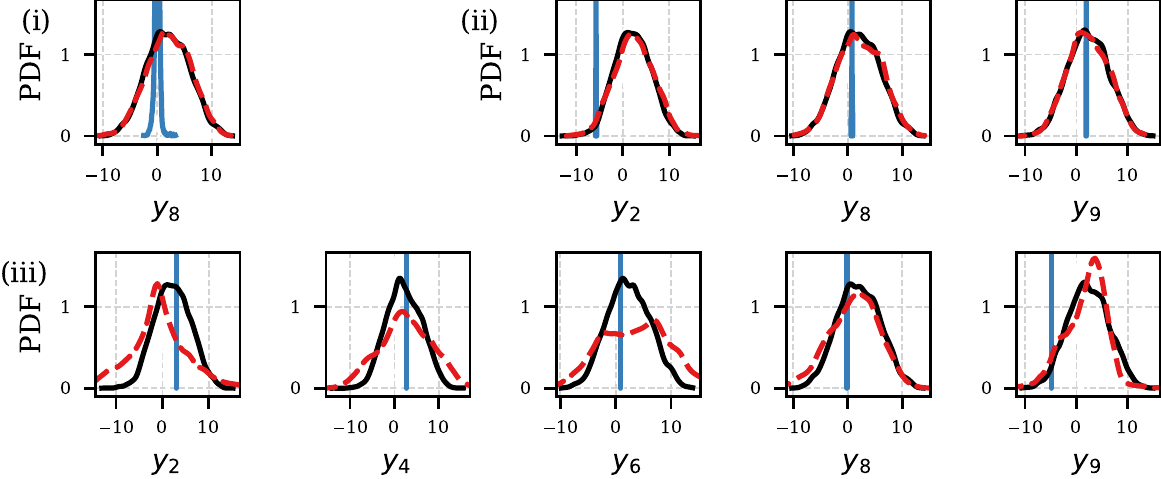}
    \caption{Statistics reconstruction of unmeasured variables. Comparison of the target (black line), PI-LSTM (red dashed line) and data-driven LSTM (blue line) probability density functions (PDF) of (i) $N_{\xi}=1$, (ii) $N_{\xi}=3$, (iii) $N_{\xi}=5$ unmeasured variables over a $1000 \tau_{\lambda}$ trajectory in closed-loop configuration.}
    \label{fig:pdf}
\end{figure}

In Fig.~\ref{fig:pdf} we show the statistics of the reconstructed variables, which are unseen during the training and based on the autonomous evolution of the network. The data-driven LSTM fails to reproduce the solution, in particular, the corresponding delta-like distribution (in blue) indicates a fixed-point. In test cases (i) and (ii), the PI-LSTM accurately reproduces the long-term behaviour of the dynamical system. At each time step, it successfully extrapolates from the partial input to the full state. Case (iii) shows that, by increasing the number of unmeasured observations, the complexity of the reconstruction is increased, and the accuracy of the reconstruction is decreased. However, the PI-LSTM provides a markedly more accurate statistical reconstruction of the target compared to the data-driven LSTM. The PI-LSTM predicts the observed variables well (not shown here), which indicates that incorporating knowledge of the underlying physics enables accurate long-term prediction of the full state.
\begin{figure}[h]
    \includegraphics{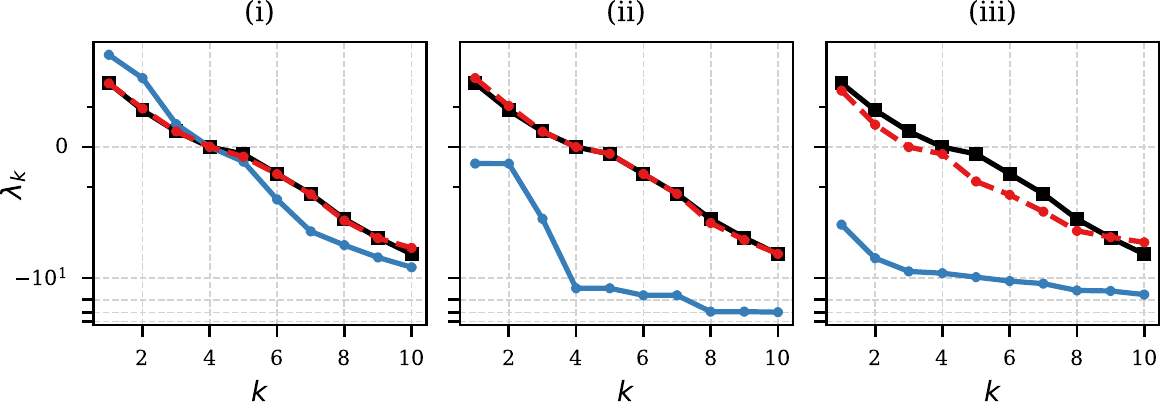}
    \caption{Comparison of the target (black squares), PI-LSTM (red dots) and data-driven LSTM (blue dots) LEs for (i) $N_{\xi}=1$, (ii) $N_{\xi}=3$, (iii) $N_{\xi}=5$ unmeasured variables. All the vertical axes are in logarithmic scale. }
    \label{fig:lyap}
\end{figure}

In Fig.~\ref{fig:lyap}, we compare the reference LEs (in black squares) with LEs extracted from the data-driven LSTM (in blue circles) and PI-LSTM (in red circles) in the three test cases. By reconstructing the unmeasured variables the networks effectively reconstruct the tangent space, the properties of which are encapsulated in the LEs. In all cases, the LEs of the data-driven LSTM deviate significantly from the reference LEs, differing more from the target when fewer observations are available. For test cases (i) and (ii), the PI-LSTM reproduces the target LEs with high accuracy, with an error of $0.28\%$ and $8\%$ in $\lambda_1$, respectively. When reducing the number of observations further, as in (iii), the accuracy of the PI-LSTM LEs is limited by the lack of information, with an error of $11.6\%$ in $\lambda_1$. Figure \ref{fig:lyap} also shows that in cases (ii) and (iii) of the data-driven LSTM, the leading LE is negative ($\lambda_1<0$), resulting in a completely incorrect solution (fixed point solution). 
This means that in (ii) and (iii) the data-driven LSTM displays no chaotic dynamics, whereas the PI-LSTM reproduces the chaotic behaviour. 




\section{Conclusions and future directions}\label{sec:conclusion}
We propose the Physics-Informed Long Short-Term Memory (PI-LSTM) network to embed the knowledge of the governing equations into an LSTM, by using the network's prediction to compute a first-order derivative. In contrast to physics-informed neural networks, which have no internal recurrent connections, the PI-LSTMs capture temporal dynamics whilst penalizing predictions that violate the system’s governing equations. We deploy the PI-LSTM to reconstruct the unmeasured variables of the Lorenz-96 system, which is a chaotic system with three positive Lyapunov exponents. The long-term prediction of the PI-LSTM in closed-loop accurately reconstructs the statistics of multiple unmeasured variables. By computing the Lyapunov exponents of the PI-LSTM, we show the key role of the physics-informed loss in learning the dynamics. This exemplifies how leveraging knowledge of the physical system can be advantageous to reconstruct and predict data in line with the fundamental chaotic properties. Future work will focus on reconstructing unmeasured variables from experimental data.

\bibliographystyle{splncs04}
\bibliography{ref}

\begin{thebibliography}{10}
\providecommand{\url}[1]{\texttt{#1}}
\providecommand{\urlprefix}{URL }
\providecommand{\doi}[1]{https://doi.org/#1}

\bibitem{doan_piesn}
Doan, N., Polifke, W., Magri, L.: Physics-informed echo state networks. Journal
  of Computational Science  \textbf{47},  101237 (2020).
  \doi{10.1016/j.jocs.2020.101237}

\bibitem{Eckmann_Ruelle1985}
Eckmann, J.P., Ruelle, D.: Ergodic theory of chaos and strange attractors.
  Reviews of Modern Physics  \textbf{57},  617--656 (1985).
  \doi{10.1103/RevModPhys.57.617}

\bibitem{FUNAHASHI_nn_approx}
Funahashi, K.I.: On the approximate realization of continuous mappings by
  neural networks. Neural Networks  \textbf{2}(3),  183--192 (1989).
  \doi{10.1016/0893-6080(89)90003-8}

\bibitem{hornik1989multilayer}
Hornik, K., Stinchcombe, M., White, H.: Multilayer feedforward networks are
  universal approximators. Neural networks  \textbf{2}(5),  359--366 (1989).
  \doi{10.1016/0893-6080(89)90020-8}

\bibitem{adam}
Kingma, D.P., Ba, J.: Adam: {A} method for stochastic optimization. In: 3rd
  International Conference on Learning Representations, {ICLR} 2015, San Diego,
  CA, USA, May 7-9, 2015, Conference Track Proceedings (2015).
  \doi{10.48550/ARXIV.1412.6980}

\bibitem{lagaris}
Lagaris, I., Likas, A., Fotiadis, D.: Artificial neural networks for solving
  ordinary and partial differential equations. IEEE Transactions on Neural
  Networks  \textbf{9}(5),  987--1000 (1998). \doi{10.1109/72.712178}

\bibitem{lorenz96}
Lorenz, E.N.: Predictability: a problem partly solved. In: Seminar on
  Predictability, 4-8 September 1995. vol.~1, pp. 1--18. ECMWF, Shinfield Park,
  Reading (1996). \doi{10.1017/CBO9780511617652.004}

\bibitem{georgios_stability_analysis}
Margazoglou, G., Magri, L.: Stability analysis of chaotic systems from data
  (2022). \doi{10.48550/ARXIV.2210.06167}

\bibitem{Pathak_ml_le}
Pathak, J., Lu, Z., Hunt, B.R., Girvan, M., Ott, E.: Using machine learning to
  replicate chaotic attractors and calculate lyapunov exponents from data.
  Chaos: An Interdisciplinary Journal of Nonlinear Science  \textbf{27}(12),
  121102 (2017). \doi{10.1063/1.5010300}

\bibitem{racca2021automatic}
Racca, A., Magri, L.: Automatic-differentiated physics-informed echo state
  network (api-esn). In: International Conference on Computational Science. pp.
  323--329. Springer (2021). \doi{10.48550/ARXIV.2101.00002}

\bibitem{raissi2019physics}
Raissi, M., Perdikaris, P., Karniadakis, G.E.: Physics-informed neural
  networks: A deep learning framework for solving forward and inverse problems
  involving nonlinear partial differential equations. Journal of Computational
  Physics  \textbf{378},  686--707 (2019). \doi{10.1016/j.jcp.2018.10.045}

\bibitem{Sangiorgio_2021}
Sangiorgio, M., Dercole, F., Guariso, G.: Forecasting of noisy chaotic systems
  with deep neural networks. Chaos, Solitons \& Fractals  \textbf{153},  111570
  (2021). \doi{10.1016/j.chaos.2021.111570}

\bibitem{Vlachas_backprop}
Vlachas, P., Pathak, J., Hunt, B., Sapsis, T., Girvan, M., Ott, E.,
  Koumoutsakos, P.: Backpropagation algorithms and reservoir computing in
  recurrent neural networks for the forecasting of complex spatiotemporal
  dynamics. Neural Networks  \textbf{126},  191--217 (2020).
  \doi{10.1016/j.neunet.2020.02.016}

\end{thebibliography}
\end{document}